\documentclass{article}
\usepackage{ijcai23}

\usepackage{times}
\usepackage{soul}
\usepackage{url}
\usepackage[hidelinks]{hyperref}
\usepackage[utf8]{inputenc}
\usepackage[small]{caption}
\usepackage{graphicx}
\usepackage{amsmath}
\usepackage{amsthm}
\usepackage{booktabs}
\usepackage{algorithm}
\usepackage{algorithmic}
\usepackage[switch]{lineno}

\usepackage{amsmath,amssymb} 
\usepackage{color}
\usepackage{booktabs}
\usepackage{nicefrac}
\usepackage{microtype}      
\usepackage{xcolor}         
\usepackage{enumitem}
\usepackage{epsfig}
\usepackage{comment}
\usepackage{array}
\usepackage{pifont}
\usepackage{graphbox}
\usepackage{epstopdf}
\usepackage{sidecap}
\usepackage{wrapfig}
\usepackage{bm}
\usepackage{multirow}
\usepackage{wrapfig}
\newcommand{\yc}[0]{\textcolor[rgb]{0.1,0.5,0.1}}

\title{Universal Adaptive Data Augmentation}

\author{
Xiaogang Xu$^1$
\and
Hengshuang Zhao$^2$
\affiliations
$^1$Zhejiang Lab\\
$^2$The University of Hong Kong\\
\emails
xgxu@zhejianglab.com, \quad
hszhao@cs.hku.hk
}

\begin{document}

\maketitle

\begin{abstract}
Existing automatic data augmentation (DA) methods either ignore updating DA's parameters according to the target model's state during training or adopt update strategies that are not effective enough.
In this work, we design a novel data augmentation strategy called ``Universal Adaptive Data Augmentation" (UADA).
Different from existing methods, UADA would adaptively update DA's parameters according to the target model's gradient information during training: given a pre-defined set of DA operations, we randomly decide types and magnitudes of DA operations for every data batch during training, and adaptively update DA's parameters along the gradient direction of the loss concerning DA's parameters.
In this way, UADA can increase the training loss of the target networks, and the target networks would learn features from harder samples to improve the generalization.
Moreover, UADA is very general and can be utilized in numerous tasks, e.g., image classification, semantic segmentation and object detection.
Extensive experiments with various models are conducted on CIFAR-10, CIFAR-100, ImageNet, tiny-ImageNet, Cityscapes, and VOC07+12 to prove the significant performance improvements brought by UADA.
\end{abstract}

\section{Introduction}
The performance of deep neural networks (DNNs) would be improved significantly when more data is available.
However, the collection of datasets is expensive. To this, a massive amount of data augmentation (DA) methods have been proposed to remedy the deficiency of supervised data.
The generalization of DNNs could be prominently improved if an effective DA method is adapted.
The current DA methods can be divided into two categories. The first category is the human-designed DA, e.g., random crop and Cutout~\cite{devries2017improved}. 
Recently, the human-designed DA methods have been gradually replaced by automatic DA approaches~\cite{zhang2019adversarial,wu2020generalization}.

\begin{table}[t]
	\centering
	\large
	\resizebox{1.0\linewidth}{!}{
		\begin{tabular}{l|lp{2.0cm}<{\centering}p{2.5cm}<{\centering}}
			\toprule[1pt]
			Task & Model & Baseline(\%)&Ours(\%)\\
			\hline
			\multirow{4}{2.3cm}{Cls. on CIFAR10}&ResNet18 &95.22 &97.08{\yc{(+1.86)}} \\
			& WideResNet-28-10 &96.14 &97.83{\yc{(+1.69)}} \\
			& Shake-Shake (26 2x32d)&96.35 &97.68{\yc{(+1.33)}} \\
			& Shake-Shake (26 2x96d)&96.96 &98.27{\yc{(+1.31)}} \\
			\hline
			\multirow{4}{2.5cm}{Cls. on CIFAR100}&ResNet18 &77.01 &79.88{\yc{(+2.87)}} \\
			& WideResNet-28-10 &81.03 &83.17{\yc{(+2.14)}} \\
			& Shake-Shake (26 2x32d)& 79.02&82.28{\yc{(+3.26)}} \\
			& Shake-Shake (26 2x96d)&82.08 &85.88{\yc{(+3.80)}} \\
			\hline
			\multirow{2}{2.5cm}{Cls. on tiny-ImageNet}& ResNet18 &63.13 &67.41{\yc{(+4.28)}} \\
			& ResNet50 &65.52 &70.81{\yc{(+5.29)}} \\
			\hline
			Seg. on Cityscapes & PSPNet50& 76.11&77.80{\yc{(+1.69)}} \\
			\hline
			Det. on VOC07+12& RFBNet& 80.10&81.00{\yc{(+0.90)}} \\
			\bottomrule[1pt]
	\end{tabular}}
	\caption{This table summarizes results of UADA on different models, datasets and tasks. For image classification (Cls.), we report the top-1 accuracy, for semantic segmentation (Seg.), we report the mIoU, for object detection (Det.), we report the mAP. ``baseline" is the most common DA strategy on the corresponding datasets.}
	\label{tab:teaser}
\end{table}

\begin{figure}[t]
	\begin{center} 
		\includegraphics[width=1.0\linewidth]{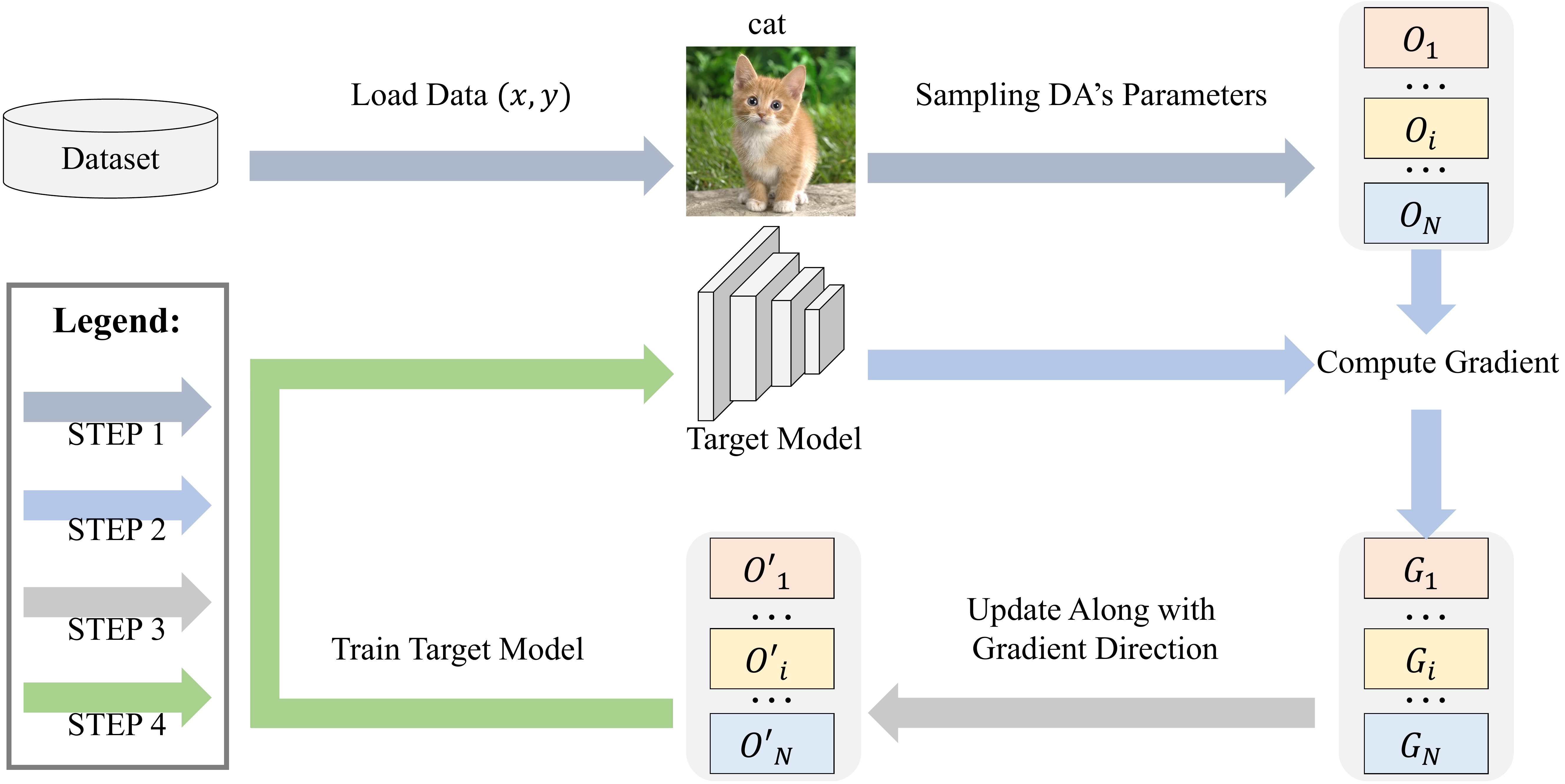}
	\end{center}
	\caption{The summary of our method's pipeline, utilizing target model's gradient to adaptively update DA's parameters during training. Details can be viewed in Fig.~\ref{fig:framework}. 
	}
	\label{fig:teaser}
\end{figure}

Existing automatic DA approaches can be divided into two main categories.
The first one would set a search phase before training to obtain DA strategies, while such methods ignore updating DA's parameters according to the target model's state during training.
The second one would optimize DA strategies during training.
However, they all adopt a common strategy which is not effective enough: randomly sampling DA's parameters and deciding the optimized parameters according to the loss value~\cite{zhang2019adversarial,wu2020generalization,lin2019online}.
On the contrary, we claim that DA's parameters should be optimized according to the target model's state, especially the gradient information.
We can compute the target model's gradient with respect to DA's parameters, and update DA's parameters along with the direction of such gradient. This strategy can effectively help to explore the hard sample, improving the target model's generalization.

In this paper, we propose a novel automatic augmentation method, called ``Universal Adaptive Data Augmentation" (UADA). Given a pre-defined set of DA operations, UADA would determine types and magnitudes of DA operations via random sampling during training, and adaptively update DA's parameters according to the target model's gradient at each training step.
With UADA, the augmented data can lead to a larger loss value, avoiding the overfit on the training data and improving the generalizability on validation data.

UADA for adaptively optimizing DA's parameters consists of four stages (as shown in Fig.~\ref{fig:teaser}):
1) randomly acquire a set of DA's parameters, load a batch of data, augment the data for training, and compute the loss value; 2) obtain the gradients of the loss with respect to the DA's parameters; 3) update DA's parameters along with the directions of the gradients; 4) utilize the optimized DA's parameters to augment the data for training, and employ the loss value and backward operation to update the weights of the network.
The tremendous challenge is the discrete property of DA's parameters, i.e., it is challenging to compute the accurate gradient of the loss concerning DA's parameters.
In contrast to backward the gradient from the network to DA's parameters, we design an effective approach to simulate the required gradients.
This gradient simulation module is motivated by the black-box adversarial method~\cite{chen2017zoo}: negligible perturbations are added to DA's parameters, and we measure the change of loss to decide the value of gradients.
In this way, an approximate value for the gradient of the loss concerning DA's parameters can be obtained with two network forwarding operations.

Our UADA is generally applicable to different tasks, including image classification, semantic segmentation and object detection, since UADA only requires knowing what kinds of DA operations will be utilized during the training.
Extensive experiments are conducted with existing representative image classification datasets, including CIFAR-10/CIFAR-100~\cite{krizhevsky2009learning}, and ImageNet~\cite{deng2009imagenet}/tiny-ImageNet~\cite{le2015tiny}.
The results on all datasets with miscellaneous network structures demonstrate the effectiveness of UADA. 
Moreover, experiments are also conducted in the semantic segmentation and object detection task, and the results on the Cityscapes~\cite{cordts2016cityscapes} and VOC07+12~\cite{pascal-voc-2012} datasets prove the effects of UADA, as summarized in Table~\ref{tab:teaser}.

In summary, our contributions in this paper are threefold.
\begin{itemize}
	\item We propose a novel Universal Adaptive Data Augmentation (UADA) strategy. This strategy can adaptively optimize DA's parameters along with the directions of loss gradient concerning DA's parameters during training, enhancing the generalization of the trained models.
	\item An effective gradient simulation approach is proposed in UADA to acquire the gradient of the loss with respect to DA’s parameters.
	\item We conduct experiments with various model structures on different datasets and tasks, which manifest the effectiveness and generality of our UADA.
\end{itemize}

\section{Related Work}

\subsection{Human-designed DA}
Existing DA methods can be divided into human-designed DA and automatic DA.
They can generate extra samples by some label-preserved transformations to increase the size of datasets and improve the generalization of networks.
Generally speaking, human-designed DA policies are specified for some datasets. Therefore, varying human-designed DA methods should be chosen for different datasets.
For example, Cutout~\cite{devries2017improved} is widely utilized in the training of CIFAR-10/CIFAR-100~\cite{krizhevsky2009learning}, while it is not suitable for ImageNet~\cite{deng2009imagenet} since Cutout would destroy the property of original samples.

\subsection{Automatic DA}
\label{sec:au-re}
AutoAugment~\cite{cubuk2019autoaugment} is the first automatic augmentation method.
The strategy searched by AutoAugment~\cite{cubuk2019autoaugment} includes the operation of ShearX/Y, TranslateX/Y, Rotate, AutoContrast, Invert, Equalize, Solarize, Posterize, Contrast, Color, Brightness, Sharpness, Cutout, and Sample Pairing.
The range of the magnitude for each operation is also discretized uniformly into ten values.
For every dataset, an augmentation policy in AutoAugment is composed of 5 sub-policies, and 5 best policies are concatenated to form a single policy with 25 sub-policies.
Each sub-policy contains two operations to be applied orderly, and each operation in AutoAugment is conducted with a chosen magnitude and a direction. 
However, it has large search space and the expensive search cost~\cite{cubuk2019autoaugment}.
Two kinds of following works are proposed: reducing the cost of search phase or searching during training to eliminate search phase.

\noindent\textbf{Reduce search space.} Recently, more and more researchers focus on reducing search space to acquire automatically-learned DA and speeding up the search process, e.g., the RandAugment~\cite{cubuk2020randaugment}, Fast AutoAugment~\cite{lim2019fast}, Faster AutoAugment~\cite{hataya2020faster}, DADA~\cite{li2020differentiable}, UA~\cite{lingchen2020uniformaugment}, and TrivialAugment~\cite{muller2021trivialaugment}.
RandAugment~\cite{cubuk2020randaugment} proposed a simplified search space that vastly reduces the computational expense of automatic augmentation; Fast AutoAugment~\cite{lim2019fast} found effective augmentation policies via a more efficient search strategy based on density matching; Faster AutoAugment~\cite{hataya2020faster} designed a differentiable policy search pipeline for data augmentation; DADA~\cite{li2020differentiable} also formulated a differentiable automatic data augmentation strategy to dramatically reduces the search cost; TrivialAugment~\cite{muller2021trivialaugment} proposed a simple strategy which is parameter-free and only applies a single augmentation to each image.

\begin{figure}[t]
	\begin{center} 
		\includegraphics[width=1.0\linewidth]{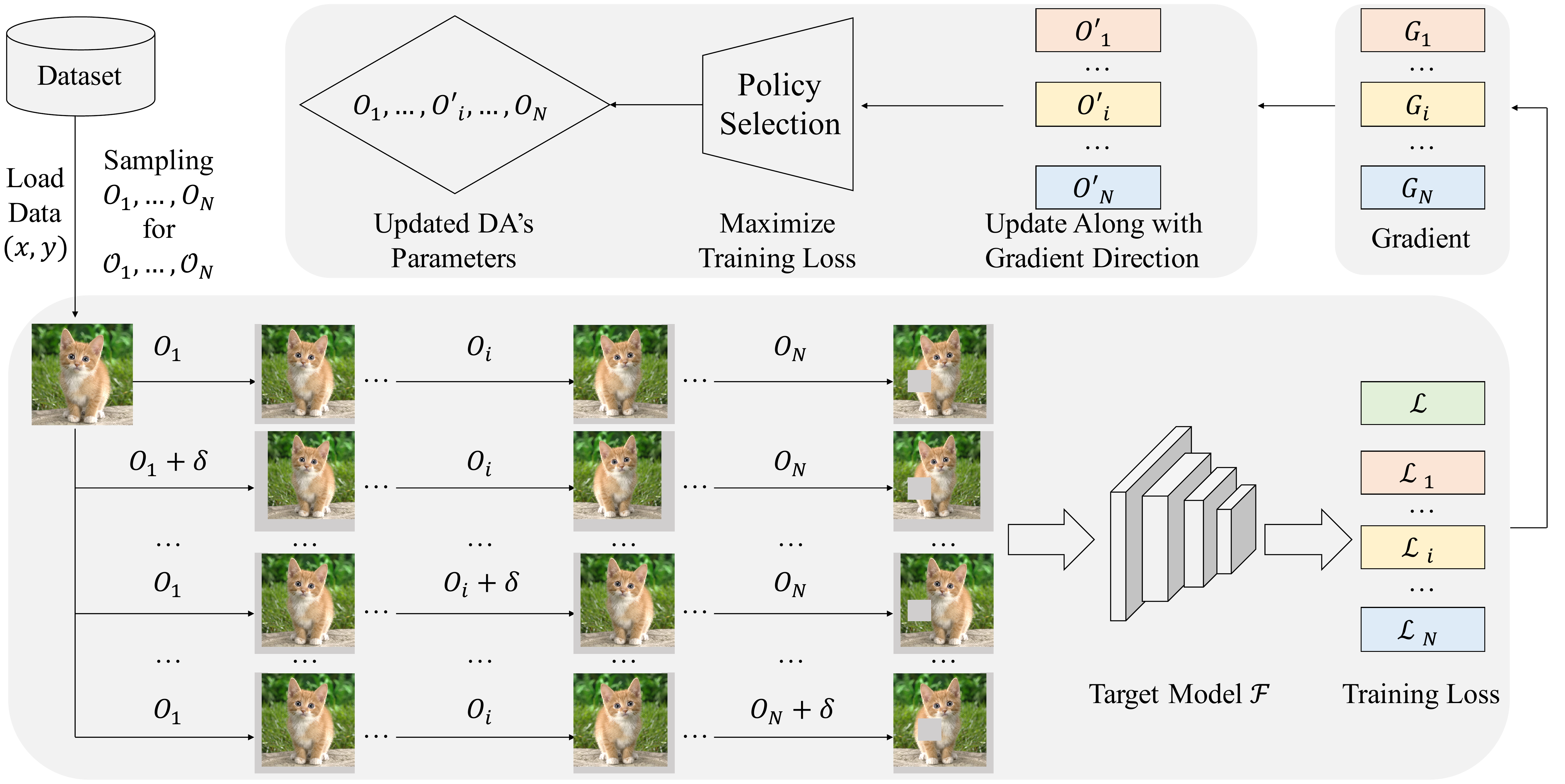}
	\end{center}
	\caption{The illustration of our Universal Adaptive Data Augmentation (UADA). Given a set of DA's parameters, we update them along with the direction of loss gradient concerning DA's parameters.
	}
	\label{fig:framework}
\end{figure}

\noindent\textbf{Search with training.} There is another kind of approach which learn the augmentation policy online as the training goes, e.g., PBA~\cite{ho2019population}, OHL~\cite{lin2019online}, Adversarial AutoAugment~\cite{zhang2019adversarial}, LTDA~\cite{wu2020generalization}, DDAS~\cite{liu2021direct}, DivAug~\cite{liu2021divaug} and OnlineAugment~\cite{tang2020onlineaugment}.
PBA~\cite{ho2019population} employed multiple workers that each uses a different policy and are updated in an evolutionary fashion; OHL~\cite{lin2019online} adopted multiple parallel workers with reinforcement learning, and defined the accuracy on held-out data after a part of training; Adversarial AutoAugment~\cite{zhang2019adversarial} made a reinforcement-learning based update, rewarding the policy yielding the lowest accuracy training accuracy, and causing the policy distribution to shift towards progressively stronger augmentations; LTDA~\cite{wu2020generalization} proposed strategy to maximize the training loss via multiple sampling for DA's parameters; DDAS~\cite{liu2021direct} exploited meta-learning with one-step gradient update and continuous relaxation to the expected training loss for efficient search; DivAug~\cite{liu2021divaug} designed a strategy to maximize the diversity of training data and hence strengthen the regularization effect.
OnlineAugment~\cite{tang2020onlineaugment} employs an augmentation model to perform data augmentation, which is differentiable and can be optimized during training. And the augmentation model can only conduct certain types of DA operations.

\noindent\textbf{Our adaptive strategy.} Different from these approaches, we propose a new framework (UADA) to online learn the augmentation policy: employ only one worker for sampling DA' parameters in each batch during training and adaptively update the DA' parameters according to the model state.
In this strategy, we randomly determine the DA operations for every data batch, randomly decide types and magnitudes of DA operations during training, and adaptively update DA's parameters according to model state at each step.

Our method is very different from existing differentiable and adversarial DA methods: 1) We directly compute the gradient of the loss concerning DA's parameters which is different from existing differentiable policy~\cite{hataya2020faster,li2020differentiable,liu2021direct}. And our strategy to obtain the gradient is achieved with adversarial learning that is different from current approaches.
2) We directly update DA's parameters along the gradient direction of loss with respect to DA's parameters, maximizing the training loss, and this is different from Adversarial AutoAugment~\cite{zhang2019adversarial}, which training a policy network with reinforcement-learning, and LTDA~\cite{wu2020generalization}, which randomly samples several sets of values for augmentation operations to conduct augmentation and choose the loss with maximum value for backward.
UADA can be utilized for various tasks, and is implemented without search strategies, avoiding the expensive search costs.

\section{Method}
\label{sec:method}

\subsection{Motivation}
Generally speaking, the target of a DA method is to avoid overfitting. To this, DA methods would normally adopt the policy to increase the training loss of the target networks, promoting the learning from hard samples.

Given a set of pre-defined DA operations, 
most of the current automatic DA methods~\cite{cubuk2019autoaugment,cubuk2020randaugment} adopt the strategy of random sampling to decide the corresponding parameters.
In contrast to prior works, we claim that the DA's parameters should be decided according to the target model's state.
Therefore, we propose to adaptively update the DA's parameters along with the direction of loss gradient concerning DA's parameters.

\begin{algorithm*}[t]
	\caption{The training algorithm with our Adaptive Adversarial Data Augmentation ($\dag$ means can be done \textit{parallel} )}
	\label{alg3}
	\begin{algorithmic}[1]
		\STATE \textit{Input:} training data $(X,Y)$, the target model $\mathcal{F}$,
		perturbation value $\delta$, step size $\epsilon$, total training epoch $t_{max}$
		\FOR {$t= 1$ to $t_{max}$}
		\STATE Load a batch of data $(x,y)$ from the training data $(X,Y)$.
		\STATE $\dag$Determine parameters of $\mathcal{O}_1,...,\mathcal{O}_N$, as $O_1,...,O_N$, via random sampling.
		\STATE $\dag$Apply the augmentation operations $\mathcal{O}_1,...,\mathcal{O}_N$ for $(x,y)$. Compute the loss $\mathcal{L}(\mathcal{F}(\mathcal{O}_N(...\mathcal{O}_1(x|O_1)|O_N)),y)$.
		\STATE $\dag$For $i=1:N$, add the perturbation $\delta$ to DA's parameters and update them with $\epsilon$ with Eq. \ref{eq:update}, obtaining $O'_1,...,O'_N$.
		\STATE $\dag$For $i=1:N$, apply the augmentation with parameters as $O_1,...O'_i,...,O_N$ for $(x,y)$ to compute the loss.
		\STATE Choose the loss with the maximal value to update the network $\mathcal{F}$.
		\ENDFOR
		\RETURN the trained network $\mathcal{F}$
	\end{algorithmic}
\end{algorithm*}

\subsection{Notation}
Given a set of DA operation $\mathcal{O}$, it consists of several operations $\mathcal{O}_i, i\in[1,N]$, where $N$ is the number of the operation.
$\mathcal{O}_i$ is combined with the corresponding parameters $O_i$, including types and magnitudes.
Assume $x$ and $y$ are the notations of data and its corresponding ground truth, respectively.
The target model for training is represented as $\mathcal{F}$, and the loss function to train the model is denoted $\mathcal{L}()$.
Associate with the chosen DA, for a batch of data $x$, the loss function can be
\begin{equation}
	\small
	\mathcal{L}(\mathcal{F}(\mathcal{O}_N(...\mathcal{O}_1(x|O_1)|O_N)),y),
\end{equation}
where $\mathcal{O}_i(\quad|O_i)$ means applying the $i$-th DA's operation, $\mathcal{O}_i$, to process the data, with its parameters $O_i$.

\subsection{The Pipeline of UADA}

\noindent\textbf{Overview.}
The overview of our UADA strategy can be viewed in Fig. \ref{fig:framework}, which consists of four stages. 
First, we load a batch of data $(x,y)$, acquire a set of DA's parameters $O_i, i\in[1,N]$ via random sampling, apply them to process the data $x$, and obtain the corresponding loss value $\mathcal{L}$. Next, we add perturbations $\delta$ to DA's parameters, augment the data, compute the loss value $\mathcal{L}_i, i\in[1,N]$, and measure the change of loss compared with $\mathcal{L}$. In this way, the gradients of the loss with respect to DA's parameters ($G_i, i\in[1,N]$) can be simulated.
Finally, we update DA's parameters with the direction of the computed gradients (maximizing the training loss), apply the updated DA's operations to process the batch of data, use the processed data to update the network's weights.
Details will be described in the following.

\noindent\textbf{Random sampling DA's parameters.}
Like normal DA strategy~\cite{cubuk2019autoaugment,cubuk2020randaugment}, we shall first obtain a set of DA's parameters $O_i, i\in[1,N]$ via random sampling.
And we can then update the corresponding values according to target model's state.
This is different from~\cite{zhang2019adversarial,wu2020generalization,lin2019online} since they would adopt multiple sampling operations to obtain multiple sets of DA's parameters during training.

\noindent\textbf{Compute gradient.}
It is nearly impossible to obtain the gradient of loss concerning DA's parameters, since DA's parameters are often non-differentiable since their values are discrete, e.g., the coordinates in the Cutout, as well as the policy of the searched AutoAugment~\cite{cubuk2019autoaugment}.
Even some operations can be differentiable, e.g., random rotation and random translation~\cite{riba2020kornia}, they would become non-differentiable if they are combined with the other non-differentiable operations.

Suppose a negligible random perturbation $\delta$ is added to the DA’s parameter $O_i$, then we can obtain intermediate DA's parameters $\widehat{O}_i=O_i+\delta$.
Meanwhile, we can compute the new loss as $\mathcal{L}(\mathcal{F}(\mathcal{O}_N(...\mathcal{O}_i(...\mathcal{O}_1(x|O_1)|\widehat{O}_i)|O_N)),y)$.
Comparing the new loss with the original loss 
$\mathcal{L}(\mathcal{F}(\mathcal{O}_N(...\mathcal{O}_i(...\mathcal{O}_1(x|O_1)|O_i)|O_N)),y)$,
we can acquire the simulated gradient of the loss with respect to the DA's parameter $O_i$, as 
\begin{equation}
	\small
	\begin{split}
		G_i=[&\mathcal{L}(\mathcal{F}(\mathcal{O}_N(...\mathcal{O}_i(...\mathcal{O}_1(x|O_1)|\widehat{O}_i)|O_N)),y)-\\ &\mathcal{L}(\mathcal{F}(\mathcal{O}_N(...\mathcal{O}_i(...\mathcal{O}_1(x|O_1)|O_i)|O_N)),y)] / \delta.
	\end{split}
	\label{eq:grad}
\end{equation}
$\delta$ is small enough (its value can be positive or negative).

\noindent\textbf{Update DA's parameters along with gradient direction.}
The adversarial learning methods~\cite{goodfellow2014explaining,madry2017towards,chen2017zoo,guo2019simple} would compute the gradient of the loss with respect to the input data $x$. The computed gradient is employed to create the perturbation and obtain the perturbed input data $x'$. A representative process, which is called FGSM~\cite{goodfellow2014explaining}, can be written as
\begin{equation}
	\small
	x'=x+\alpha \cdot sign(\bigtriangledown_{x} (\mathcal{L}(\mathcal{F}(x), y))),
\end{equation}
where $\bigtriangledown_{x}$ means computing the gradient concerning the input data $x$, $sign()$ is the operation to return the sign of the input data, and $\alpha$ is the step size, i.e., the magnitude of the perturbation.
The perturbed data $x'$ can cause an increase of loss for the target model $\mathcal{F}$.
Training with both clean samples $x$ and the perturbed data $x'$ can enhance the robustness of the target model $\mathcal{F}$. However, such training will also sacrifice the generalization of the target model, leading to a decrease in performance on clean samples in the validation set.

In this paper, we apply the idea of adversarial learning to the update of DA's parameters, resulting in the enhancement of the target model's generalization.
Given a batch of data $x$, we compute the loss as $\mathcal{L}(\mathcal{F}(\mathcal{O}_N(...\mathcal{O}_1(x|O_1)|O_N)),y)$. Suppose we can compute the gradient of the loss with respect to the DA's parameters, the parameters of each DA's operation can be updated as
\begin{equation}
	\resizebox{0.7\linewidth}{!}{$O'_i=O_i+\epsilon \cdot sign(\bigtriangledown_{O_i} (\mathcal{L}(\mathcal{F}(\mathcal{O}_N(...\mathcal{O}_1(x|O_1)|O_N)),y)))$},
\end{equation}
where $\bigtriangledown_{O_i} (\mathcal{L}(\mathcal{F}(\mathcal{O}_N(...\mathcal{O}_1(x|O_1)|O_N)),y))$ is indeed $G_i$ in Eq.~\ref{eq:grad}.
The training with DA's parameters $O'_i$ can also lead to the increase of loss compared with the training with $O_i$. 

Thus, the updating of the parameter $O_i$ can be written as
\begin{equation}
	\small
	\begin{split}
		\widehat{O}_i &=O_i+\delta\\
		G_i&=[\mathcal{L}(\mathcal{F}(\mathcal{O}_N(...\mathcal{O}_i(...\mathcal{O}_1(x|O_1)|\widehat{O}_i)|O_N)),y)\\
		&-\mathcal{L}(\mathcal{F}(\mathcal{O}_N(...\mathcal{O}_i(...\mathcal{O}_1(x|O_1)|O_i)|O_N)),y)] / \delta,\\
		O'_i&=O_i+\epsilon \cdot sign(G_i),
	\end{split}
	\label{eq:update}
\end{equation}
where $\epsilon$ is the positive hyper-parameter for control, and it should be small enough for the accuracy of simulated gradients.
During the training, there are usually multiple DA's operations. 
We select the operation that has the most decisive influence on the change of loss value, i.e., its gradient is crucial enough. 
The details of the training algorithm by using UADA can be viewed in Alg. \ref{alg3}.

\section{Experiments}

\subsection{Datasets}
For the classification task, the datasets include CIFAR-10/CIFAR-100~\cite{krizhevsky2009learning}, and ImageNet~\cite{deng2009imagenet}/tiny-ImageNet~\cite{le2015tiny}.
For experiments of image classification, we report the accuracy value (\%).
For the semantic segmentation and object detection, experiments are conducted on Cityscapes~\cite{cordts2016cityscapes} and VOC07+12~\cite{pascal-voc-2012}.

\subsection{Implementation Details}
\label{sec:detail}
We use PyTorch~\cite{paszke2017automatic} to implement ResNet-18~\cite{he2016identity}, WideResNet-28-10~\cite{zagoruyko2016wide}, and Shake-Shake~\cite{gastaldi2017shake} for CIFAR-10/CIFAR-100; ResNet-18 and ResNet-50 for ImageNet~\cite{deng2009imagenet}/tiny-ImageNet~\cite{le2015tiny}; PSPNet~\cite{zhao2017pyramid} for the Cityscapes~\cite{cordts2016cityscapes}; SSD~\cite{liu2016ssd} and RFBNet~\cite{liu2018receptive} for VOC07+12~\cite{pascal-voc-2012}.

$\delta$ is set as 1, and $\epsilon$ in Eq. \ref{eq:update} is also set as 1 in the classification experiments (except the ablation study).
And the values of ($\delta$, $\epsilon$) could be decided for different datasets adaptively according to the augmentation types.
To ensure the accuracy of the estimated gradient, the value of ($\delta$, $\epsilon$) should not be large. Thus, the search space for the hyper-parameter ($\delta$, $\epsilon$) is not large for different datasets.
And to decide the optimal value of ($\delta$, $\epsilon$) for a target dataset, the value range of ($\delta$, $\epsilon$) can be quantized into several points.
The search can be completed on these points without much tuning.
Especially, as proved by our ablation study, the performance of UADA is not sensitive to the value of ($\delta$, $\epsilon$).

\begin{table}[t]
	\centering
	\Huge
	\resizebox{1.0\linewidth}{!}{
		\begin{tabular}{l|cccccccc}
			\toprule[1pt]
			CIFAR-10&Baseline& Cutout& AA & RA & DADA & Fast AA&Faster AA & PBA \\
			\hline
			ResNet18 &95.22& 96.17 &96.49&96.33 & -& -&- &- \\
			WideResNet-28-10&96.14& 96.96&97.34&97.47 &97.30 &97.30 &97.40 & 97.40 \\
			Shake-Shake (26 2x32d)&96.35&96.90 &97.53&97.38&97.30&97.30&97.30&97.46 \\
			Shake-Shake (26 2x96d)&96.96&97.44 &97.98&98.03&98.00&98.00&98.00&97.97 \\
			\hline
			CIFAR-10&O.A.&TA&UA& LTDA & DDAS & DivAug& Adv. AA &\textbf{Ours}\\
			\hline
			ResNet18 &-&-&-&-& -&- &- &\textbf{97.08} \\
			WideResNet-28-10&97.60&97.46& 97.33&97.89& 97.30&\textbf{98.10} &97.50 &97.83 \\
			Shake-Shake (26 2x32d)&-&-&-&-&- & - &97.64 &\textbf{97.68} \\
			Shake-Shake (26 2x96d)&-&98.21&98.10&98.22&97.90 &98.10  & 98.15& \textbf{98.27}\\
			\bottomrule[1pt]

			\toprule[1pt]
			CIFAR-100&Baseline&Cutout& AA& RA & DADA & Fast AA&Faster AA&PBA\\
			\hline
			ResNet18 &77.01& 77.50 &79.05&78.75 & -&-  &- &- \\
			Shake-Shake (26 2x32d)&79.02&80.55  &82.20&82.03 & -&- &- &- \\
			Shake-Shake (26 2x96d)&82.08& 83.08 &85.61&85.12 &84.70 &85.10 &84.40&84.69 \\
			\hline
			CIFAR-100&O.A.&TA&UA& LTDA& DDAS& DivAug& Adv. AA &\textbf{Ours}\\
			\hline
			ResNet18 &-&-&-&-&- & -&- &\textbf{79.88} \\
			Shake-Shake (26 2x32d)&-&-&-&-&- &- & -&\textbf{82.28} \\
			Shake-Shake (26 2x96d)&-&85.16&85.00&-&84.90 & 85.30 &\textbf{85.90} &85.88 \\
			\bottomrule[1pt]
	\end{tabular}}
	\caption{The comparison with SOTA DA methods on CIFAR.}
	\label{tab:sota}
\end{table}

\begin{table}[t]
	\centering
	\Large
	\resizebox{1.0\linewidth}{!}{
		\begin{tabular}{l|p{2.1cm}<{\centering}p{2.1cm}<{\centering}p{2.1cm}<{\centering}p{2.1cm}<{\centering}p{2.1cm}<{\centering}}
			\toprule[1pt]
			Method & RA&Adv. AA& LTDA& DivAug& UADA\\
			\hline
			Training($\times$) &1.0&8.0 &6.0&4.5 &4.2 \\
			\bottomrule[1pt]
		\end{tabular}
	}
	\caption{Comparing the training cost among our method, RA, Adv. AA, LTDA and DivAug on CIFAR-10 relative to RA.}
	\label{tab:comparison2-time}
\end{table}

\begin{table}[t]
	\centering
	\Large
	\resizebox{1.0\linewidth}{!}{
		\begin{tabular}{l|p{2.5cm}<{\centering}p{2.5cm}<{\centering}|p{2.5cm}<{\centering}p{2.5cm}<{\centering}}
			\toprule[1pt]
			\multirow{2}{*}{Setting} & \multicolumn{2}{c|}{ResNet18} &\multicolumn{2}{c}{ResNet50}\\
			\cline{2-5}
			&top1 & top5 &top1 & top5\\
			\hline
			Baseline&69.82&89.32&76.79&93.40 \\
			\hline
			Ours&\textbf{70.50}&\textbf{89.51}&\textbf{77.19}&\textbf{93.50} \\
			\bottomrule[1pt]
	\end{tabular}}
	\caption{Experiments on ImageNet with UADA.}
	\label{tab:comparison22}
\end{table}

\begin{table}[t]
	\centering
	\Large
	\resizebox{1.0\linewidth}{!}{
		\begin{tabular}{l|p{5.5cm}<{\centering}p{5.5cm}<{\centering}}
			\toprule[1pt]
			Setting & ResNet18 &ResNet50\\
			\hline
			Baseline &63.13&65.52\\
			AA&64.94&67.59\\
			RA&66.41&68.64\\
			\hline
			\textbf{Ours}&\textbf{67.41}
			&\textbf{70.81}\\
			\bottomrule[1pt]
	\end{tabular}}
	\caption{Experiments on tiny-ImageNet with UADA.}
	\label{tab:comparison2}
\end{table}

\begin{table}[t]
	\centering
	\huge
	\resizebox{1.0\linewidth}{!}{
		\begin{tabular}{l|l|p{3.0cm}<{\centering}p{4.5cm}<{\centering}}
			\toprule[1pt]
			Dataset & Setting & ResNet18&WideResNet28-10\\
			\hline
			\multirow{3}{3.0cm}{CIFAR-10}&random update strategy&96.31&97.28\\
			&update to minimize loss&89.24&92.23\\
			&our update strategy&\textbf{97.08}&\textbf{97.83}\\
			\hline
			\multirow{3}{3.0cm}{CIFAR-100}&random update strategy&79.54&82.95\\
			&update to minimize loss&74.22&79.26\\
			&our update strategy&\textbf{79.88}&\textbf{83.17}\\
			\bottomrule[1pt]
	\end{tabular}}
	\caption{The results of the ablation study, evaluating the performance of alternative strategies for updating DA's parameters. We report the accuracy values.}
	\label{tab:abla}
\end{table}

\subsection{UADA for Image Classification}

\noindent\textbf{Results of CIFAR-10/CIFAR-100.}
CIFAR-10/CIFAR-100 dataset has total 60,000 images. 50,000 images are set as the training set and 10,000 images are set as the test set.
Each image has the size of $32\times 32$ belongs to one of 10 classes.
CIFAR-10/CIFAR-100 has been extensively studied with previous data augmentation methods, and we first test our proposed UADA on this data.
The results on CIFAR-10/CIFAR-100 are reported in Tables~\ref{tab:sota}. 
These results have shown that our adaptive strategy can stably outperform most of the baselines with different model structures, where ``baseline" is the setting of~\cite{devries2017improved} (randomly crop and random horizontal flip) and ``Cutout" is the combination of randomly crop and random horizontal flip and Cutout.
UADA can lead to more significant improvement for AutoAugment (AA)~\cite{cubuk2019autoaugment} and RandAugment (RA)~\cite{cubuk2020randaugment}, compared with the advancement of human-designed DA.

Moreover, the comparisons between UADA and more state-of-the-art (SOTA) approaches with complex model structures, are shown in Tables~\ref{tab:sota}. 
These top-ranking methods contain Adversarial AutoAugment (Adv. AA)~\cite{zhang2019adversarial}, PBA~\cite{ho2019population}, DADA~\cite{li2020differentiable}, Fast AutoAugment~\cite{lim2019fast} (Fast AA), Faster AutoAugment~\cite{hataya2020faster} (Faster AA), UA~\cite{lingchen2020uniformaugment}, LTDA~\cite{wu2020generalization}, DDAS~\cite{liu2021direct}, DivAug~\cite{liu2021divaug}, O.A.~\cite{tang2020onlineaugment}, and TA~\cite{muller2021trivialaugment}.
As shown in this table, the performance of our UADA is higher than most of these approaches with different network structures on CIFAR-10 and CIFAR-100.
The performance of Adversarial AutoAugment, LTDA, DivAug, and UADA on CIFAR is similar while our UADA has fewer training hours as shown in Table \ref{tab:comparison2-time}. The training cost of Adv. AA is cited from~\cite{zhang2019adversarial}, the cost of LTDA is cited from~\cite{wu2020generalization} and the cost of DivAug is cited from~\cite{liu2021divaug}.

\noindent\textbf{Results of ImageNet/tiny-ImageNet.}
We shall evaluate the effectiveness of UADA on large-scale datasets, e.g., 
ImageNet~\cite{deng2009imagenet} which is a significantly challenging dataset in image recognition.
We employ two network structures, ResNet18 and ResNet50, for experiments.
The batch size is 256, the learning rate is 0.1, the weight decay is 0.0001, the momentum is 0.9. Moreover, we adopt the cosine learning rate for training and train the network for 100 epochs.
The results are shown in Table \ref{tab:comparison22} where ``Baseline" includes RandomResizedCrop and RandomHorizontalFlip.
These results demonstrate that UADA can also lead to performance improvement on ImageNet.

Meanwhile, we also conduct experiments with a sub-set of ImageNet called tiny-ImageNet~\cite{le2015tiny}.
The tiny-ImageNet contains 100,000 images of 200 classes for training, and has 10,000 images for testing.
The results on tiny-ImageNet are reported in Table \ref{tab:comparison2}, where ``baseline" is the setting of~\cite{lee2019rethinking} including random crop and random horizontal flip.
Clearly, our UADA can also increase the performance of AA and RA.

\begin{table}[t]
	\centering
	\Large
	\resizebox{1.0\linewidth}{!}{
		\begin{tabular}{l|l|p{1.5cm}<{\centering}p{1.5cm}<{\centering}p{1.5cm}<{\centering}p{1.5cm}<{\centering}}
			\toprule[1pt]
			\multirow{2}{1.5cm}{Dataset}&\multirow{2}{1.5cm}{Model}&\multicolumn{4}{c}{$\epsilon$}\\
			\cline{3-6}
			& &0 & 1&2&3\\
			\hline
			\multirow{2}{2cm}{CIFAR-10}&ResNet18 &96.33&97.08&96.95&96.65\\
			&WideResNet28-10 &97.47&97.83&97.75&97.71\\
			\hline
			\multirow{2}{2cm}{CIFAR-100}&ResNet18 &78.75&79.88&80.07&79.25\\
			&WideResNet28-10 &82.80&83.17&83.41&83.43\\
			\bottomrule[1pt]
	\end{tabular}}
	\caption{The analysis for the influence of step size. ``$\epsilon$'' is the parameter of the step size in Eq. \ref{eq:update}, ``0'' means the results without adaption, i.e., the results of baselines.}
	\label{tab:abla2}
\end{table}

\begin{table}[t]
	\centering
	\Large
	\resizebox{1.0\linewidth}{!}{
		\begin{tabular}{l|p{1.5cm}<{\centering}p{1.5cm}<{\centering}p{1.5cm}<{\centering}|p{1.5cm}<{\centering}p{1.5cm}<{\centering}p{1.5cm}<{\centering}}
			\toprule[1pt]
			Model&\multicolumn{3}{c|}{PSPNet50}&\multicolumn{3}{c}{PSPNet101}\\
			\hline
			Setting & mIoU&mAcc &allAcc& mIoU&mAcc &allAcc\\
			\hline
			Baseline &76.11&83.80&95.77&78.22&85.65&96.12\\
			Ours&\textbf{77.80}&\textbf{84.87}&\textbf{96.02}&\textbf{78.83}&\textbf{86.08}&\textbf{96.16}\\
			\bottomrule[1pt]
	\end{tabular}}
	\caption{Experiments conducted on the semantic segmentation task, where ``baseline" is the setting containing RandScale, RandRotate, RandomGaussianBlur, RandomHorizontalFlip, and random crop.}
	\label{tab:comparison3}
	
	\resizebox{1.0\linewidth}{!}{
		\begin{tabular}{l|p{5.0cm}<{\centering}|p{5.0cm}<{\centering}}
			\multicolumn{3}{c}{}\\
			\toprule[1pt]
			Setting&SSD300(VGG16) & RFBNet300(VGG16)\\
			\hline
			Baseline &76.9 & 80.1\\
			Ours & \textbf{77.8} &  \textbf{81.0} \\
			\bottomrule[1pt]
	\end{tabular}}
	\caption{Experiments conducted on the object detection task, where ``baseline" is the original default DA setting for SSD and RFBNet. We report the mAP.}
	\label{tab:comparison33}
\end{table}

\subsection{Ablation Study}

\noindent\textbf{Alternative strategies for updating parameters.}
In our strategy, we update the DA's parameters to increase the loss value, i.e., the parameters are updated along with the direction of the computed loss gradient concerning DA parameters.
And there are several alternative adaptive strategies, including
\begin{itemize}
	\item The DA's parameters are updated with random strategy, e.g., $O'_i=O_i+\epsilon \cdot sign(R_i)$, compared with Eq. \ref{eq:update} ($R_i$ is the variable sampled from standard normal distribution).
	\item The DA's parameters are updated with the adversarial strategy that aims to minimize the loss. Different from our strategy, this alternative strategy would update DA's parameters in reverse to the direction of the gradient, e.g., $O'_i=O_i-\epsilon \cdot sign(G_i)$, compared with Eq. \ref{eq:update}.
\end{itemize}
We set the experiments to demonstrate the superiority of our adaptive strategy over these alternative strategies.
We conduct experiments with the image classification task on CIFAR-10/CIFAR-100 dataset, employing the model structures as ResNet18 and  WideResNet28-10.
The results are shown in Table \ref{tab:abla}.
Obviously, the performance of these alternative adaptive strategies is weaker than our UADA.

\noindent\textbf{The influence of step size.}
During the training, a hyper-parameter is the step size $\epsilon$  in updating DA's parameters. We will analyze the impact of step size on the final performance.
The step size is set as 1 for the experiments in the above sections, and we change its value to 2 and 3 in this experiment.
The experimental results are shown in Table \ref{tab:abla2}. Compared to results of other approaches as in Tables~\ref{tab:sota}, the change of step size will influence the final performance of the trained network while the corresponding performance is still higher than most of the baselines.

\subsection{UADA for Semantic Segmentation}
\label{sec:seg-uada}
\begin{figure}[h]
	\centering
	\newcommand\widthpose{0.35}
	\newcommand\heightpose{0.230}
	\Huge
	\resizebox{1.0\linewidth}{!}{
		\begin{tabular}{c@{\hspace{0.5mm}}c@{\hspace{0.5mm}}c@{\hspace{0.5mm}}c@{\hspace{0.5mm}}c}
			\rotatebox{90}{\parbox[t]{15mm}{\hspace*{\fill}\hspace*{\fill}}}&
			\includegraphics[width=\widthpose\textwidth, height=\heightpose\textwidth]{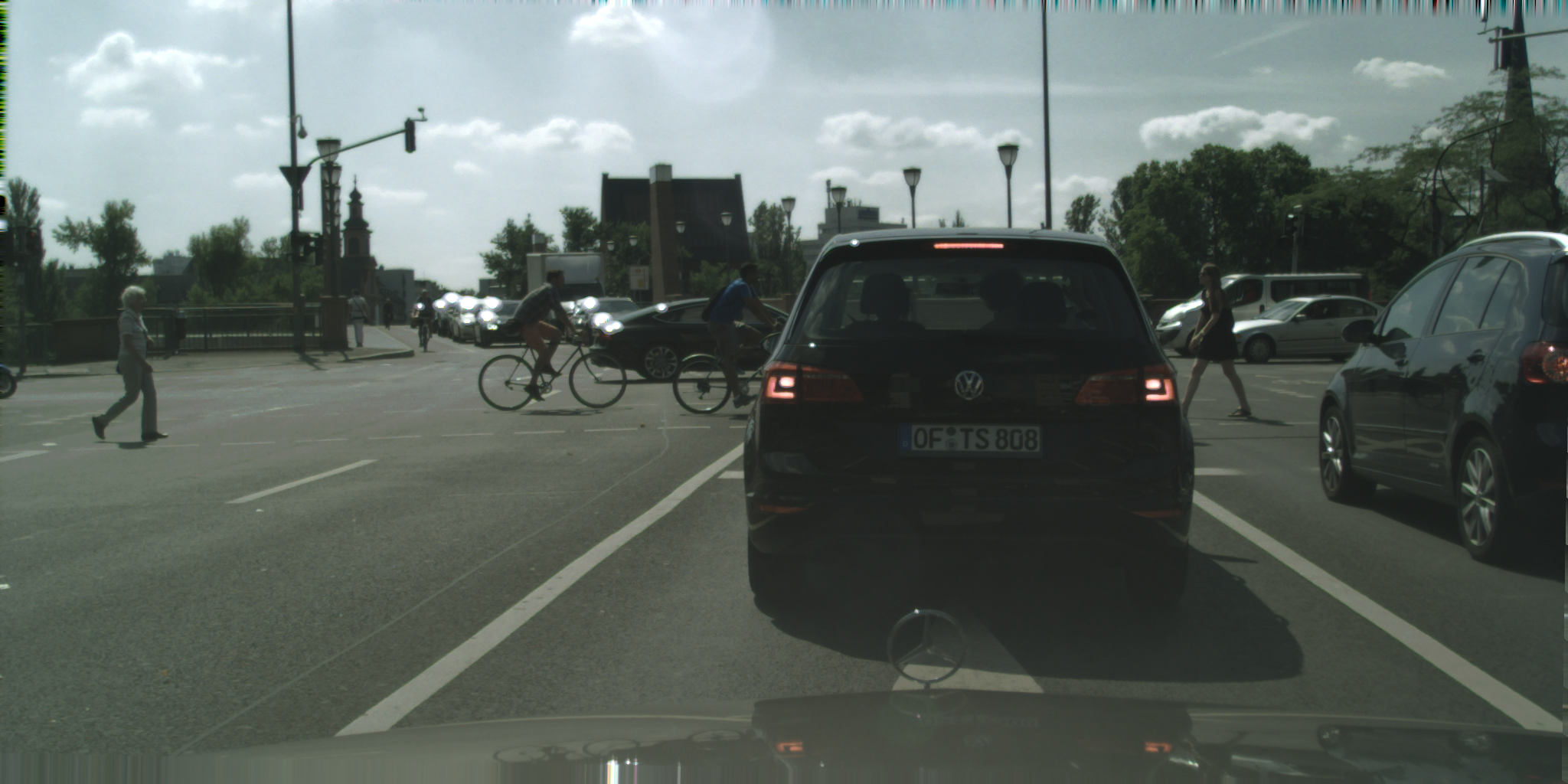} &
			\includegraphics[width=\widthpose\textwidth, height=\heightpose\textwidth]{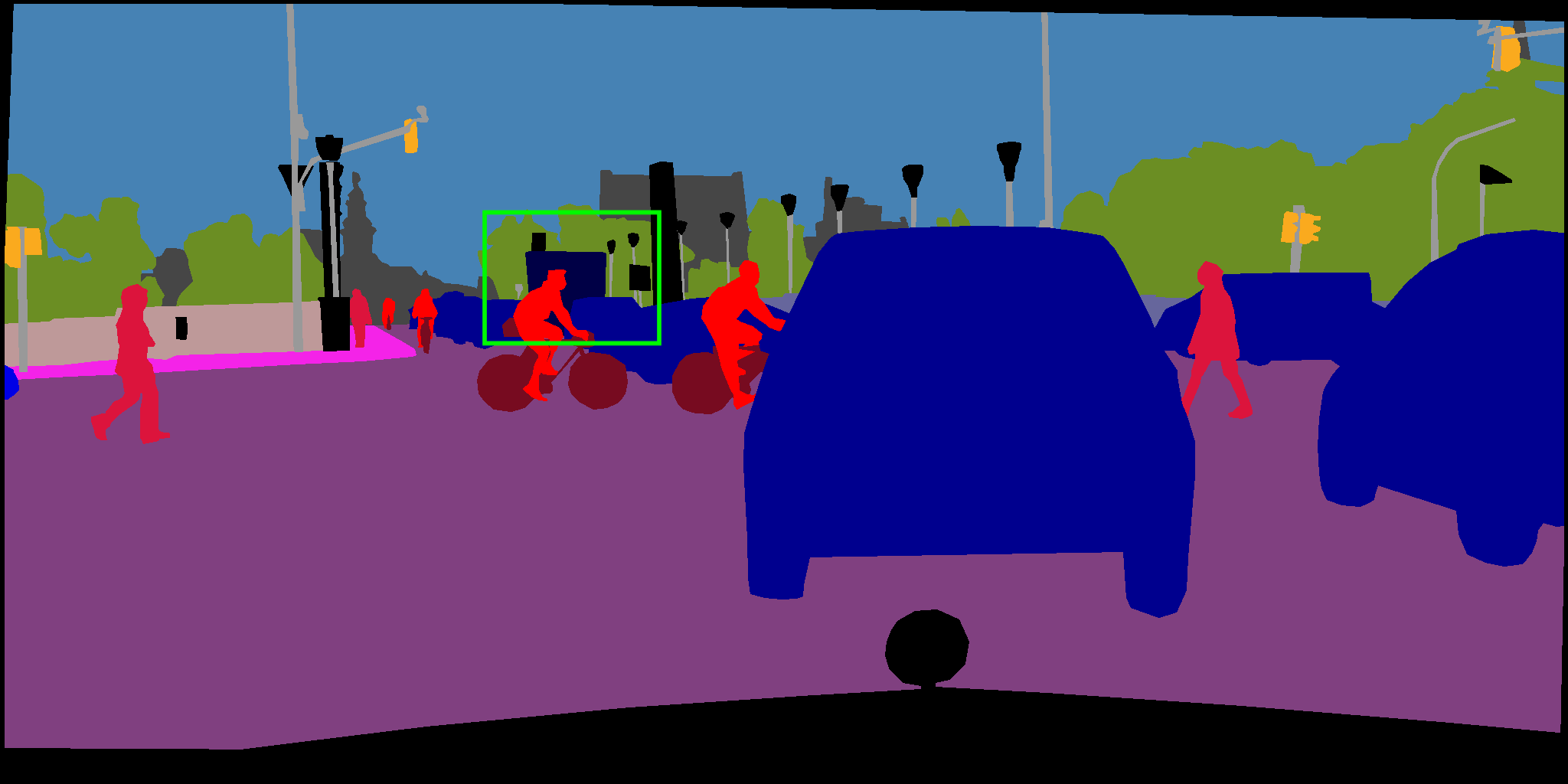}&
			\includegraphics[width=\widthpose\textwidth, height=\heightpose\textwidth]{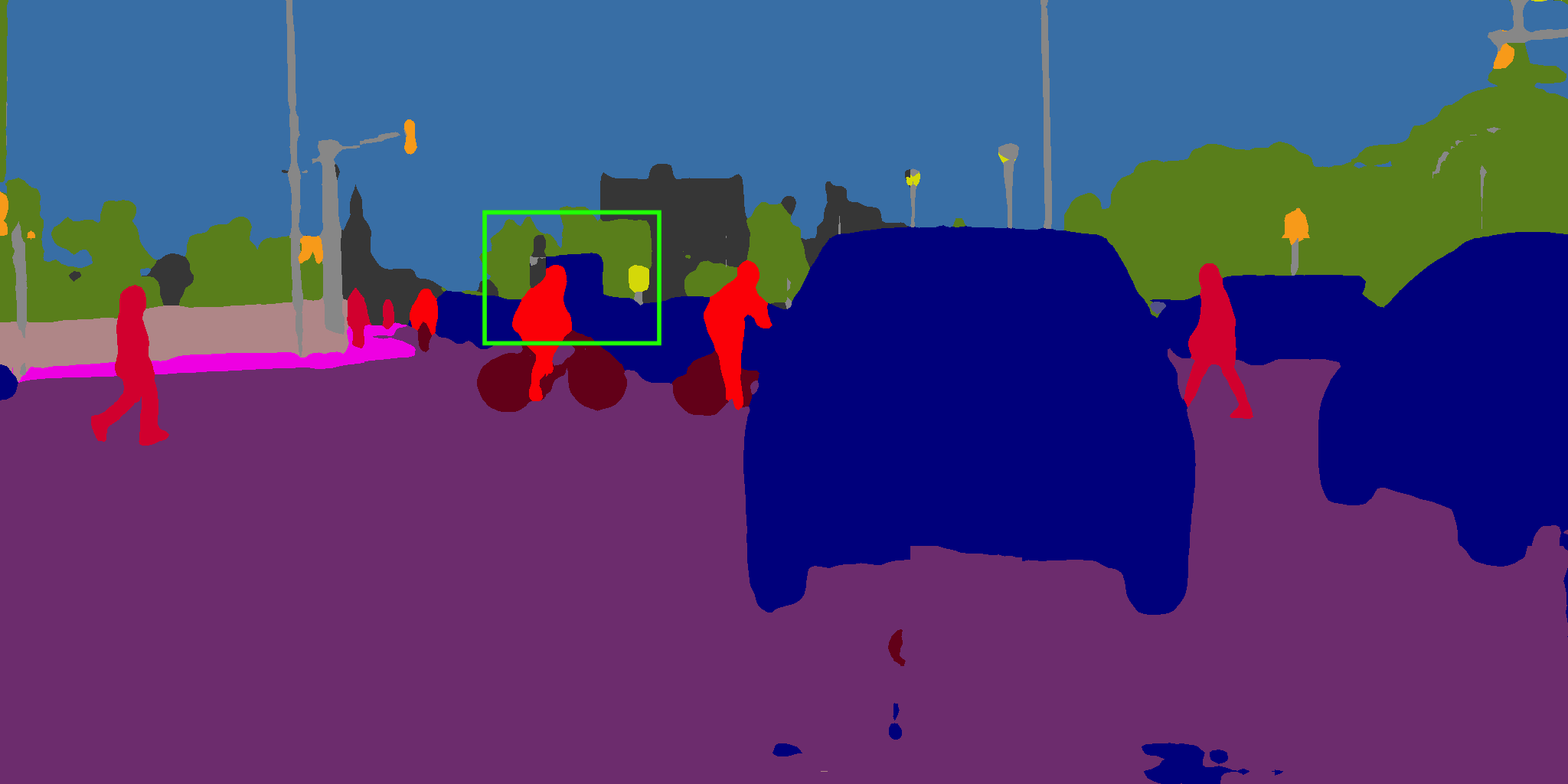} &
			\includegraphics[width=\widthpose\textwidth, height=\heightpose\textwidth]{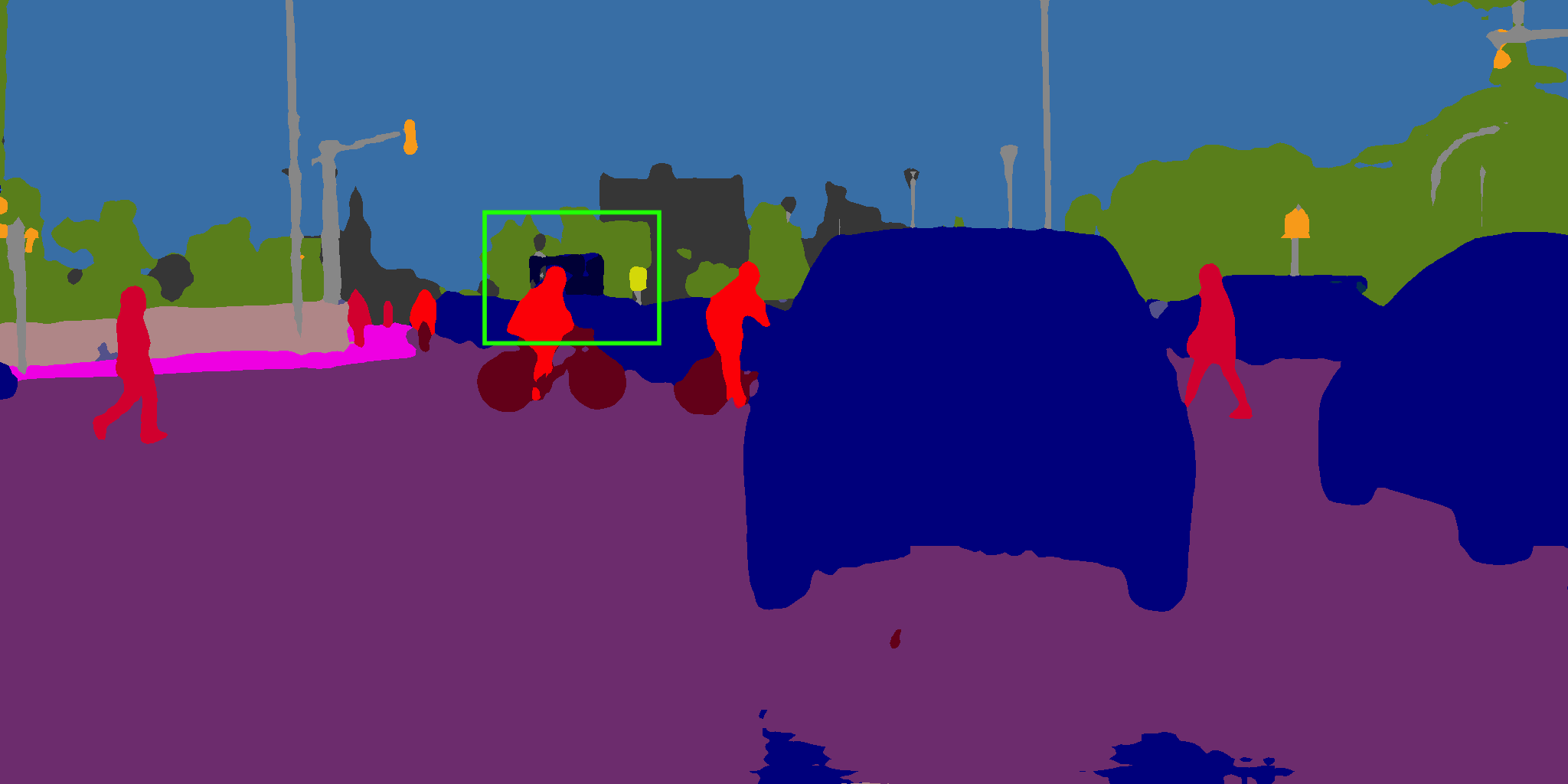}   \\
			\rotatebox{90}{\parbox[t]{15mm}{\hspace*{\fill}\hspace*{\fill}}}&
			\includegraphics[width=\widthpose\textwidth, height=\heightpose\textwidth]{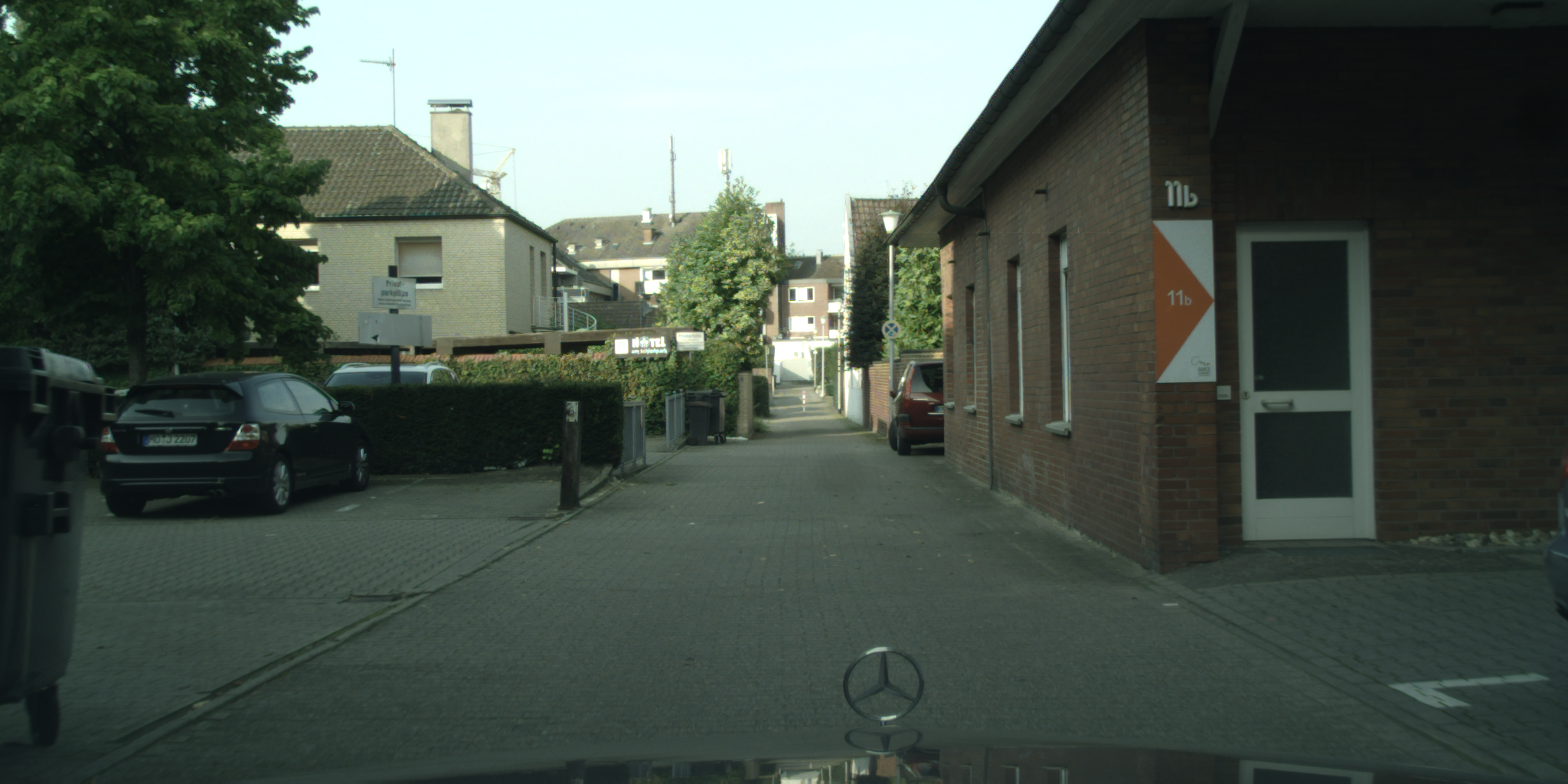} &
			\includegraphics[width=\widthpose\textwidth, height=\heightpose\textwidth]{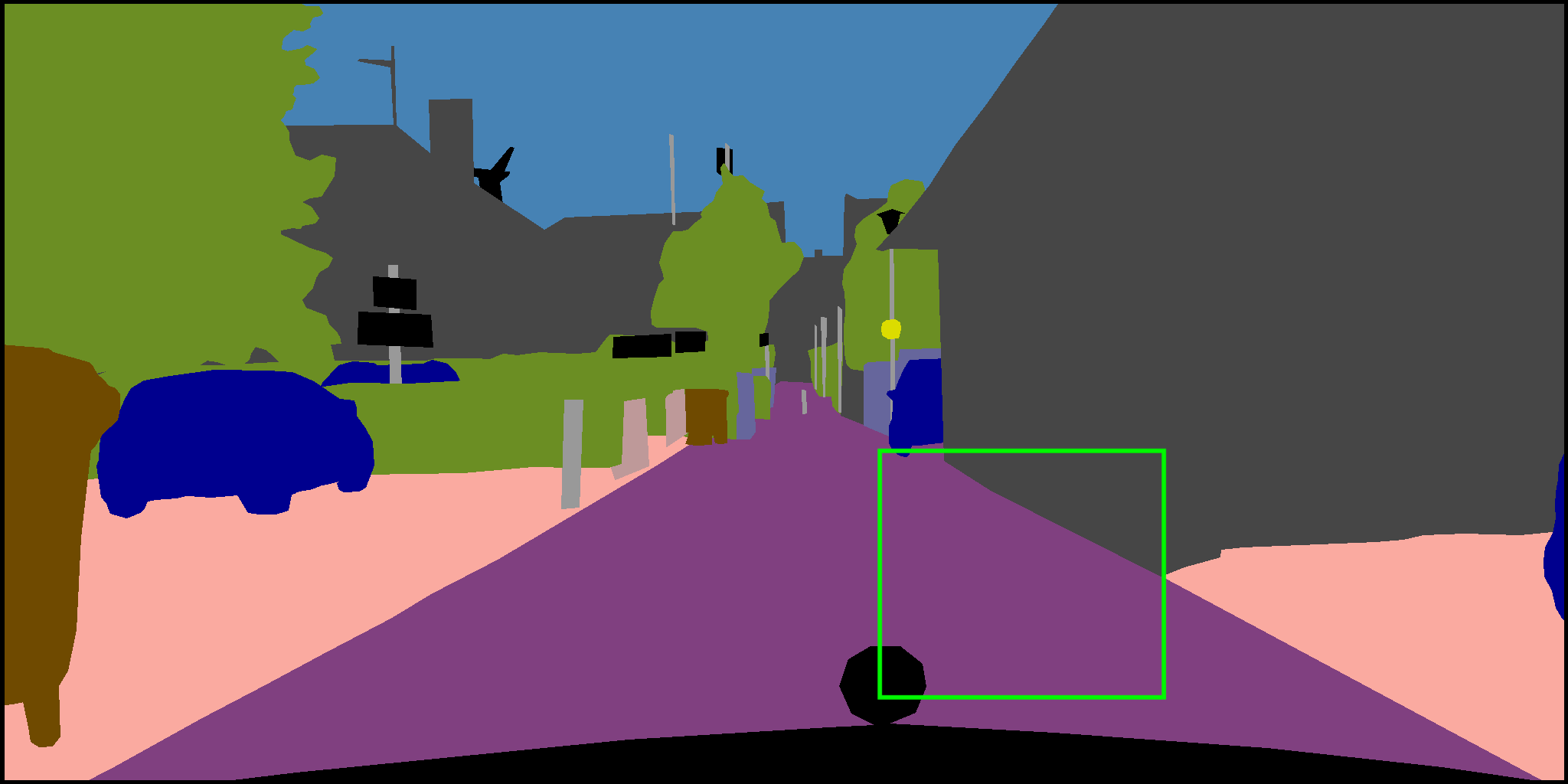}&
			\includegraphics[width=\widthpose\textwidth, height=\heightpose\textwidth]{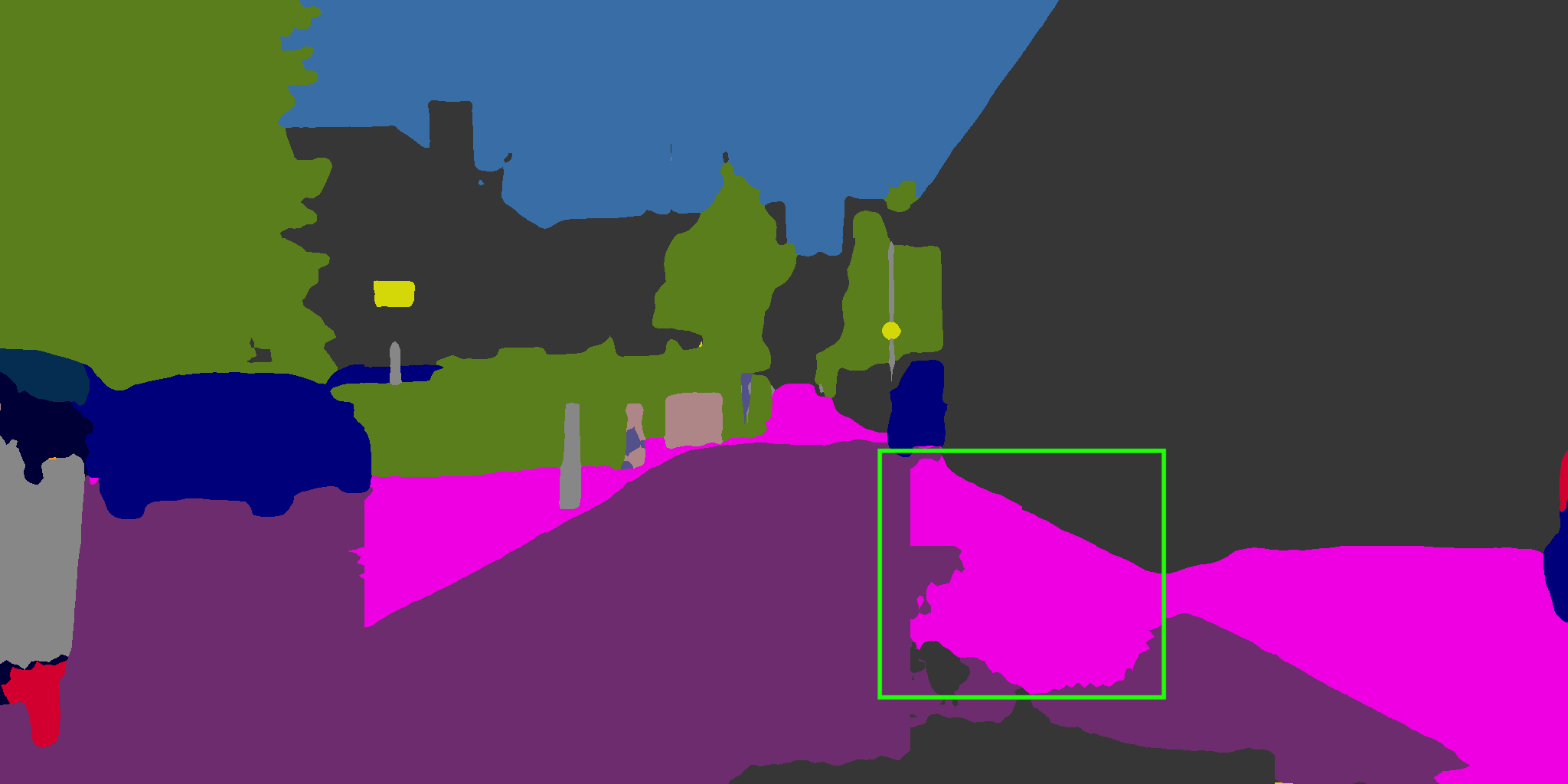} &
			\includegraphics[width=\widthpose\textwidth, height=\heightpose\textwidth]{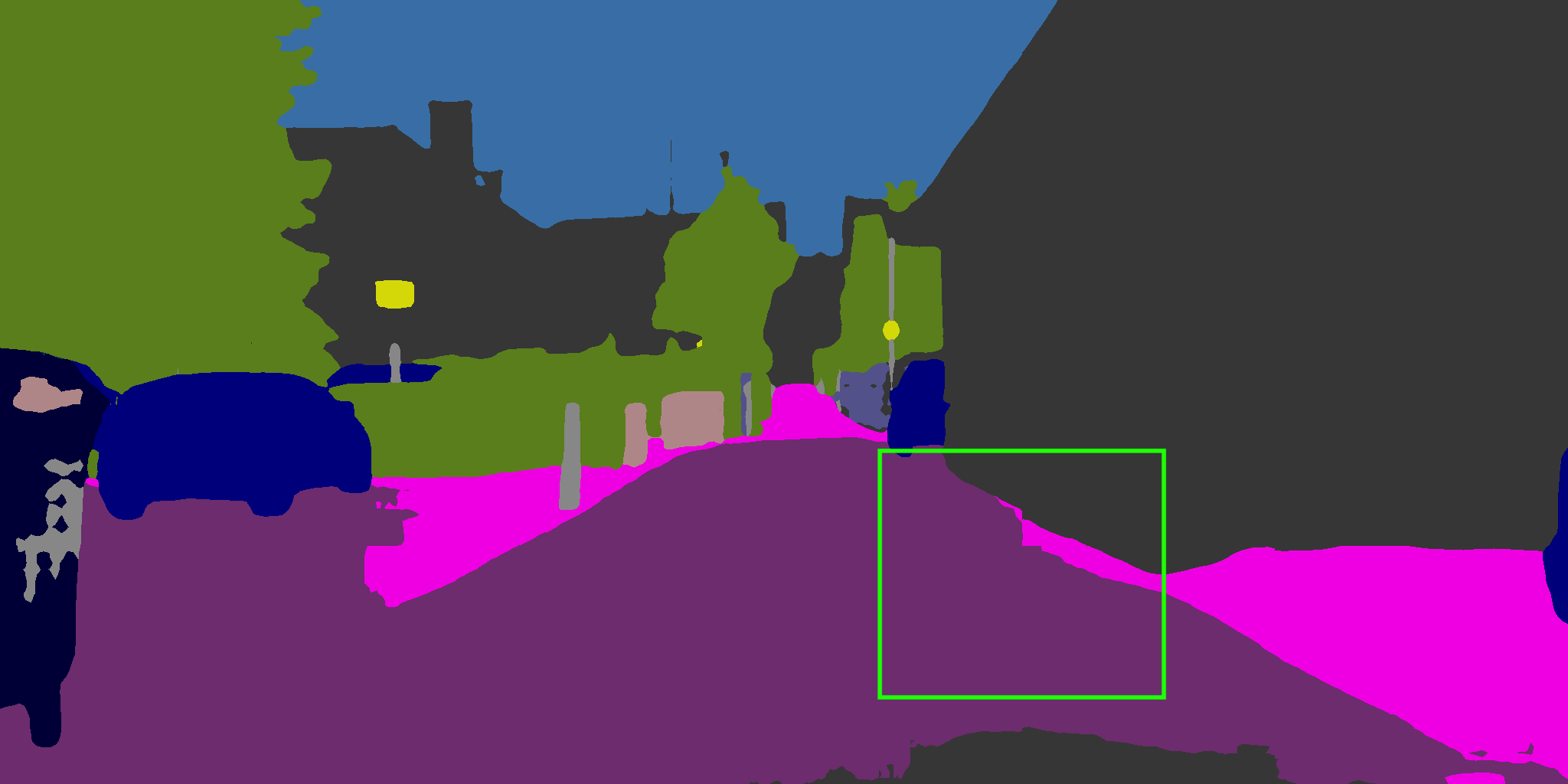} \\
			&&&&\\
			&Image & Ground Truth & Baseline & UADA \\
			
	\end{tabular}}
	\caption{Visual comparisons for different DA methods, which are executed on Cityscapes with PSPNet50 (top two rows) and PSPNet101 (bottom two rows). From the view in the yellow rectangle, we can see more clear differences.}
	\label{fig:visual_city_psp}
\end{figure}

Compared with the image classification task, the semantic segmentation task aims to give the category prediction of all pixels.
Thus, during the training, we employ the sum of all pixels' loss in one image as the loss of one image.
To demonstrate that our method can also be applied to the semantic segmentation task, we conduct the experiments on Cityscapes.
The Cityscapes dataset is collected for urban scene understanding with 19 categories. It contains high-quality pixel-level annotations with 2,975, 500, and 1,525 images for training, validation, and testing.
The network structure is set as PSPNet with different backbones, ResNet50 and ResNet101~\cite{he2016deep}.
Since existing automatic DA approaches mainly focus on image classification, there is no automatic DA method for segmentation.
Thus, we set the baseline as the DA strategy, which is commonly adopted.
The results are displayed in Table \ref{tab:comparison3} and we can observe the improvement brought by our UADA.
And visual samples are displayed in Fig. \ref{fig:visual_city_psp}.

\subsection{UADA for Object Detection}
UADA is very general for different tasks, and we apply UADA for the detection task in this section.
The experiments are conducted on VOC07+12~\cite{pascal-voc-2012}, and  we choose the detection model of SSD~\cite{liu2016ssd} and RFBNet~\cite{liu2018receptive} for experiments.
SSD is implemented by following the data augmentation strategy in \url{https://github.com/amdegroot/ssd.pytorch}, and the same DA operations are applied for the training of RFBNet.
Moreover, the training parameters are kept as the default for SSD and RFBNet.
As shown in Table~\ref{tab:comparison33}, our UADA can be applied with these DA operations, and UADA can bring noticeable performance improvement for both SSD and RFBNet.

\section{Conclusion}
In this paper, we propose a universal adaptive data augmentation approach (UADA),
where we compute the gradient of the loss with respect to the DA's parameters and use them to update the DA's parameters during training. 
With such a strategy, we can explore the hard samples during the training and avoid overfitting.
Our UADA can be applied to different tasks.
Extensive experiments are conducted on the image classification task and the semantic segmentation task with different networks and datasets, proving the effectiveness of our proposed UADA. 

In the future, we would explore more effective gradient simulation methods. Moreover, as a general data augmentation strategy, we would prove the effects of our UADA for more tasks, including middle-level (e.g., depth estimation) and low-level computer vision tasks (e.g., super-resolution and denoising).

\noindent\textbf{Relation with current DA methods.}
Besides the noticeable differences with existing DA approaches in terms of the augmentation pipeline (as carefully discussed in ``Related Work" Section), 
UADA can also be utilized to improve the results of existing DA methods: 
given the DA's parameters set by existing DA methods during training, UADA can update these parameters along the direction of loss gradient concerning DA's parameters to improve the generalization.

\appendix

\section*{Ethical Statement}

There are no ethical issues.

\section*{Acknowledgments}

This work is supported by Research Initiation Project of Zhejiang Lab (No. 2022PD0AC02).

\bibliographystyle{named}
\bibliography{ijcai23}

\end{document}